\RequirePackage{fix-cm}
\documentclass[smallextended]{svjour3}       
\smartqed  
\usepackage{graphicx}
\usepackage{multicol}
\usepackage{amsmath}
\usepackage{amsfonts}
\usepackage[labelfont=bf]{caption}
\newcommand{\ra}[1]{\renewcommand{\arraystretch}{#1}}
\usepackage{booktabs}
\usepackage[margin=0.75in]{geometry}

\usepackage[hidelinks]{hyperref} 
\newcommand{\JimSan}{Jim\'{e}nez-S\'{a}nchez}

\usepackage{etoolbox}

\usepackage{color}

\usepackage{comment}

\usepackage{xspace}

\def\etal{\textit{et al.}\xspace}
\def\ie{\textit{i.e.}~}

\def\vs{\emph{vs}.~}
 
\usepackage{float}

\usepackage[]{changes}
\definechangesauthor[name={Diana Mateus}, color=red]{DM}
\definechangesauthor[name={Sonja Kirchhoff},color=teal]{SK}
\definechangesauthor[name={Shadi Albarqouni},color=green]{SA}
\definechangesauthor[name={Anees Kazi},color=orange]{AK}
\definechangesauthor[name={Amelia},color=blue]{AJS}

\newcommand{\params}{{\bf p}}
\newcommand{\im}{{\bf I}}
\newcommand{\crop}{\im'}
\newcommand{\lab}{y}

\newcommand{\locw}{\omega_g}
\newcommand{\pred}{\hat{y}}
\newcommand{\class}{f}
\newcommand{\classw}{\omega_f}
\newcommand{\warp}{\mathcal{W}_{\params}}
\newcommand{\real}{\mathbb{R}}

\begin{document}

\title{Precise Proximal Femur Fracture Classification for Interactive Training and Surgical Planning}

\author{ Amelia \JimSan \and Anees Kazi \thanks{A. \JimSan~and A. Kazi have contributed equally to this work.} \and Shadi Albarqouni \and Chlodwig Kirchhoff \and Peter Biberthaler \and Nassir Navab \and Sonja Kirchhoff \and Diana Mateus \thanks{S. Kirchhoff and D. Mateus are joint senior authors.}
}

\institute{Amelia \JimSan, \email{amelia.jimenez@upf.edu} \at 
BCN MedTech, DTIC, Universitat Pompeu Fabra. Barcelona, Spain. \and
A. Kazi, S. Albarqouni and N. Navab \at Computer Aided Medical Procedures, Technische Universit{\"a}t M{\"u}nchen. Munich, Germany. \\ \and
S. Albarqouni is also with \at Computer Vision Lab, ETH Z{\"u}rich, Z{\"u}rich. Switzerland. \\ \and
C. Kirchhoff, P. Biberthaler, S. Kirchhoff \at
Department of Trauma Surgery, Klinikum rechts der Isar, Technische Universit{\"a}t M{\"u}nchen. Munich, Germany. \\ \and
D. Mateus \at 
Ecole Centrale de Nantes, LS2N, UMR CNRS 6004. Nantes, France.
}


\date{Received: / Accepted: date}

\maketitle

\begin{abstract}
\textbf{Purpose}. Demonstrate the feasibility of a fully automatic computer-aided diagnosis (CAD) tool, based on deep learning, that localizes and classifies proximal femur fractures on X-ray images according to the AO classification. The proposed framework aims to improve patient treatment planning and provide support for the training of trauma surgeon residents.

\textbf{Material and Methods}. A database of 1347 clinical radiographic studies was collected. Radiologists and trauma surgeons annotated all fractures with bounding boxes, and provided a classification according to the AO standard. In all experiments, the dataset was split patient-wise in three with the ratio 70\%: 10\%: 20\% to build the training, validation, and test sets, respectively. ResNet-50 and AlexNet architectures were implemented as deep learning classification and localization models, respectively. Accuracy, precision, recall and $F_1$-score were reported as classification metrics. Retrieval of similar cases was evaluated in terms of precision and recall. 

\textbf{Results}. The proposed CAD tool for the classification of radiographs into types “A”, “B” and “not-fractured”, reaches a $F_1$-score of 87\% and AUC of 0.95, when classifying fractures \textit{versus} not-fractured cases it improves up to 94\% and 0.98. Prior localization of the fracture results in an improvement with respect to full image classification. 100\% of the predicted centers of the region of interest are contained in the manually provided bounding boxes. The system retrieves on average 9 relevant images (from the same class) out of 10 cases.

\textbf{Conclusion}. Our CAD scheme localizes, detects and further classifies proximal femur fractures achieving results comparable to expert-level and state-of-the-art performance. Our auxiliary localization model was highly accurate predicting the region of interest in the radiograph. We further investigated several strategies of verification for its adoption into the daily clinical routine. A sensitivity analysis of the size of the ROI and image retrieval as a clinical use case were presented.


\end{abstract}

\section{Introduction}
\label{intro}
Proximal femur fractures are a significant problem especially of the elderly population in the western world. Starting at an age of 65 the incidence of femoral fractures increases exponentially and is almost doubled every five years. The consequences of proximal femur fractures have a significant socioeconomic impact since the mortality rate one year after the accident ranges between 14 and 36\% \cite{Ryan2015,Giannoulis2016,DeBellis2014}.

\begin{figure*}[t]
\centering
	\includegraphics[width=0.9\textwidth]{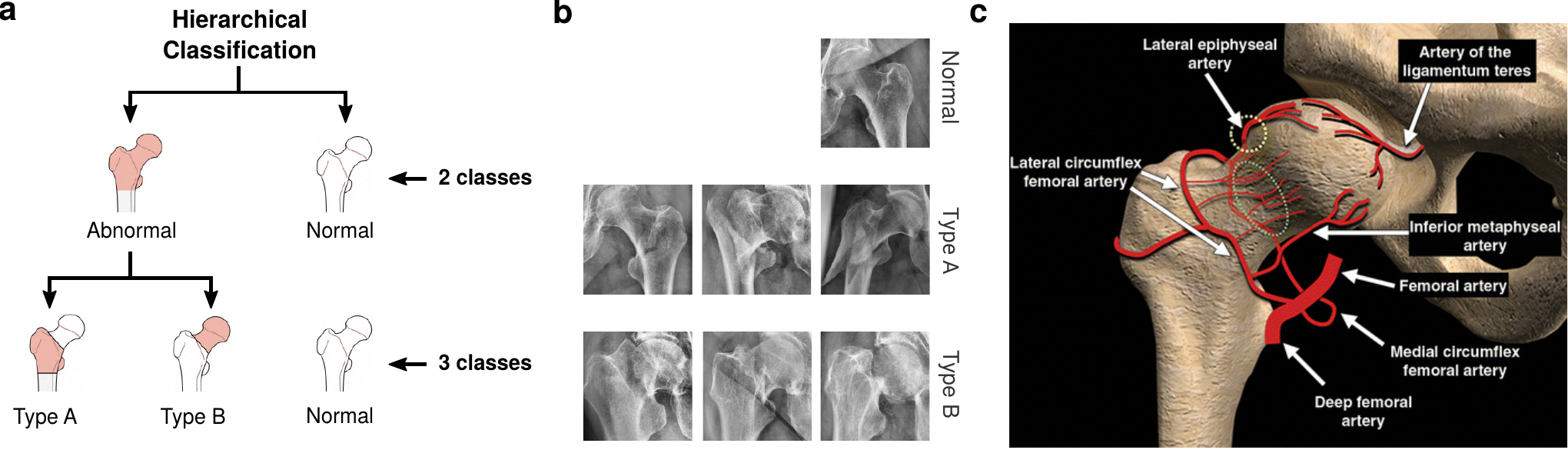}
	\caption{(a) Hierarchical classification according to the AO standard. Two scenarios are considered fracture detection (2-class), and classification of the fracture into type A or B (3-class). (b) Examples of regions of interest of X-ray images in our dataset, from top to bottom: healthy femur, fracture type A and B, respectively are shown. (c) Vascular anatomy of the proximal femur, adapted from \cite{Sheehan2015}.
	}
	\label{fig:diagram}
\end{figure*}

In almost all cases, surgical treatment has to be considered the gold standard \cite{Bhandari2015}. If surgical treatment is decelerated, several complications, as well as an increase in mortality rates, may result \cite{Zuckerman1996hip,Grigoryan2014ortho}. Early detection and classification of proximal femur fractures are crucial for the indication of surgery and, if so, to choose the adequate surgical implant. For the determination of the optimal treatment option, the vascular anatomy of the proximal femur plays an essential role, see Fig.~\ref{fig:diagram}-(c). These fractures are often described as subcapital, transcervical, or basicervical regarding its location along with displaced \textit{versus} non-displaced. This differentiation is a key factor because the blood supply to the femoral head is at risk following intraarticular femur fractures. Elderly patients suffer more frequently from transverse subcapital femur fractures arising from rather low-energy trauma compared to younger individuals ($<$ 65 years) showing a tendency to vertical distal femur neck or basicervical fractures resulting from rather high-energy trauma \cite{Swiontkowski1984fractures,Parkkari1999majority}. In this context, the Arbeitsgemeinschaft f{\"u}r Osteosynthesefragen (AO-Foundation) established a generally applicable and valid classification system for fractures of all bones of the skeleton based on X-rays \cite{Kellam2018} including the proximal femurs. 

In the literature, the AO classification for proximal femur fractures was claimed to present a better reproducibility compared to other classifications such as the Jensen \cite{Jin2005}. In cases of subcapital femur fractures in elderly patients, the Garden classification is more frequently used. However, since the Garden classification describes only subcapital femur fractures, the more extensive AO classification, which includes also intertrochanteric fractures, was used in this study. The AO classification is hierarchical, as shown in Figure~\ref{fig:diagram}-(a), and is determined by the localization and configurations of the fracture lines. In case the trochanteric region is broken, fractures are considered as “A" while those affecting the subcapital area “B" fractures. In general, the treatment is oriented to a restoration of mobility and to prevent relapse after surgery. The treatment method depends on the location of the fracture, displacement of fragments, and further concomitant patients’ facts like age and functional demands.

The skills for correctly classifying proximal femur fractures are trained during daily clinical routine in the trauma surgery department. However, the learning curve of young trauma surgery residents, especially if working in small peripheral hospitals, is long and shallow. It takes several years of practice to become an expert, as shown by the significant difference in the inter-reader agreement of 66\% among residents \vs 71\% among experienced trauma surgeons \cite{vanEmbden2010}. Currently, young trauma surgeons and medical students majorly rely on the judgment call of colleagues and attendants to achieve a correct classification to choose the best therapeutic option for the patient. Although, there are several online support systems available such as the “bone ninja" or the “AO surgical reference”, these do only demonstrate the different fracture classifications comparable to a textbook. Currently, there is no available automated system capable of classifying X-ray images individually and fracture-specifically. 

Therefore, this work aimed to develop a computer-aided diagnosis (CAD) tool based on radiographs to automatically identify proximal femur fractures in a first step, and consecutively classify them according to the AO classification. Such a CAD system can not only help in the correct classification of fractures but also be effective in planning the optimal therapy for the individual patient since the adequate treatment plan arises from the initial classification. 

In this paper, we show that Convolutional Neural Networks (ConvNets) trained on X-ray images and image-wise class-annotations constitute a suitable predictive model for automatic and on the fly classification of fractures according to the AO standard. We demonstrate the applicability of such models on a clinical dataset of 1347 radiographs. The achieved performance is similar to that of expert radiologists and trauma surgeons reported in \cite{vanEmbden2010}. 
We further propose a modification of the direct classification workflow considering a localized region of interest (ROI) around the fracture, which further improves the classification results.
Finally, we address the question of how to effectively integrate such tool into the clinical routine by 
\begin{itemize}
    \item performing a sensitivity analysis of the size of the localized ROI,
    \item investigating the potential of retrieval for the training of young trauma surgeons.
\end{itemize}

\section{Related Work}
From a technical point of view, the automated image analysis of fractures presents significant challenges due to the poor contrast and large variability of the images (see Fig.~\ref{fig:diagram}-(b)). 
Such difficulties are exacerbated for proximal femur fractures due to background clutter and the presence of overlapping structures in the pelvic region \cite{wu2012fracture}.
Initial prior work for detection and classification of fractures \cite{al2013detecting,bayram:2016diffract} focused on conventional machine learning pipelines consisting of preprocessing, feature extraction and classification steps. Predictions are based on hand-crafted features which are sensitive to the low quality of X-ray images. For example, Bayram~\etal~\cite{bayram:2016diffract} relied on the number of fragments to classify diaphyseal femur fractures. More recently, deep learning has overcome some limitations of such approaches thanks to the integration of the discriminative feature learning within the predictive models. 

The power of ConvNets for fracture detection, that is, for the binary fracture \vs not-fractured classification task, has been demonstrated for various anatomical regions, such as spine \cite{roth2016deep}, wrist \cite{olczak2017artificial}, ankle \cite{Kitamura2019}, pelvis \cite{wu2012fracture}, and hip \cite{Urakawa2019}. Badgeley~\etal{}~\cite{Badgeley2019} investigated the complementary added value of hospital process variables and patient demographics for predicting the presence of fractures comparably to radiographs alone \cite{Badgeley2019}. Another studied aspect is the pretraining of the deep models \cite{Cheng2019,Wang2019WeaklySupervised}. Most works in medical imaging use ImageNet dataset as a pretraining material \cite{Urakawa2019,Badgeley2019,Esteva2017}. Instead, Cheng~\etal{}~\cite{Cheng2019} showed that training first the model on an easier task (body part detection on radiographs), resulted in an improvement when later optimizing for the hip fracture detection. Wang~\etal{}~\cite{Wang2019WeaklySupervised} also approached the hip fracture detection employing a sequential pipeline. First, a deeper model was trained to learn high levels of abstraction for binary classification. From this pretrained model, ROIs were extracted in a weak supervised way. Then, a shallower network was trained on the mined ROIs targetting hip and pelvic fracture detection. These methods above point to the effectiveness of ConvNets to assist the radiologist's analysis, reducing the false negative rate and boosting the speed of decisions. However, all of them target still the binary detection problem (abnormal \vs not-fractured).

To the best of our knowledge, our team is the first to treat the multi-class classification problem critical for the surgical planning. We demonstrate in this paper, that the localization of a ROI is not only important for binary detection, as suggested by Wang~\etal{}~\cite{Wang2019WeaklySupervised}, but even more for the multi-class problem. Different from \cite{Wang2019WeaklySupervised} and our preliminary work \cite{kazi2017}, where weakly-supervised strategies were explored, here, we focus on the supervised case given its superior performance.
 
\section{Methods}
\label{sec:method}
\begin{figure*}[t]
\centering
	\includegraphics[width=0.7\textwidth]{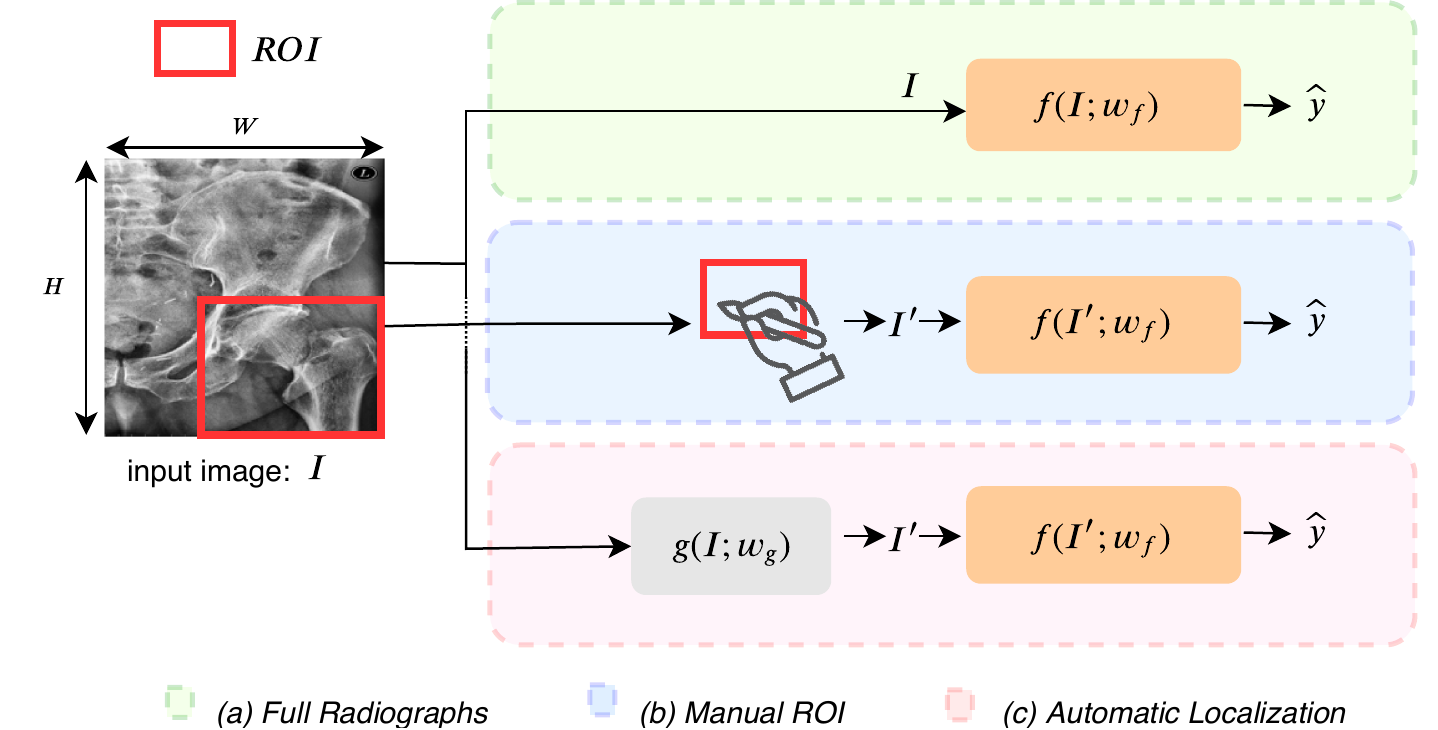}
	\caption{\textbf{Schematic representation of all the considered models. } In detail, classification on (a) Full Radiographs, (b) Manual ROIs, and (c) after Automatic Localization.}
	\label{fig:overview}
	\vspace{-.5cm}
\end{figure*}

Towards improving the clinical training and treatment planning, we aim at developing an automatic CAD system based on ConvNets capable of detecting fractures present on an X-ray image and further predicting its class according to the AO standard.
We purposely restrict the choice of architectures and optimization schemes to simple existing methods, and focus instead on developing and evaluating relevant use cases on clinical data.

In practice, given $N$ X-ray images with each image $\im \in \real^{H \times W}$, our goal is to train a {\it classification model} $\class(\cdot)$ that assigns to each image a class label $\lab \in C$, where $C$ is either $C \subset \{ \textrm{not-fractured, fracture}\}$ for detection or $C \subset \{ \textrm{not-fractured, A, B}\}$ for classification. The conceived model depends on parameters $\classw$ and provides as output a class prediction $\pred = f(\im; \classw)$.

In addition, we define an auxiliary {\it localization task} $g(\cdot)$ that returns the position $\params$ of the ROI, $\crop$, within the X-ray image such that
$\hat{\params} = g(\im; \locw)$, where $\locw$ are the localization model parameters. $\params = \{t_{r}, t_{c}, s\}$ is a bounding box of scale $s$ centered at $(t_{r}, t_{c})$. The ROI image $\crop \in \real^{H'\times W'}$ is obtained as $\crop = \warp(\im)$,
where $\warp(\cdot)$ is a warping operator. In the following, we detail three variants of the proposed CAD system to solve and combine the classification and localization tasks differently.

\subsection{Classification of Full Radiographs}
\label{ssec:AC}

We designed the baseline CAD model to receive a radiograph as input and compute the predicted class in real-time. Figure~\ref{fig:overview}-(a) shows a simplified diagram of the used CAD-principle.

Formally, the baseline model is thus $\hat{y} = f(\im; \classw)$, where $I$ stands for the whole X-ray image.
The mapping $f$ is approximated with a ConvNet optimized to minimize the cross-entropy loss function:
\begin{equation} \label{eq:wce}
\mathcal{L}_{class} = - \sum_{j\in C} y_{j,c}\log(\hat{y}_{j,c}).
\end{equation}

\subsection{Classification on Manual ROIs}
Here, we investigated the influence of localizing a relevant ROI prior to the classification. The ROI was provided by our experts, who manually drew a square containing the head and neck of the femur. We opted for a ROI around the proximal femur instead of a smaller ROI around the fracture, in order to provide contextual information. The cropped image was then used as input to the CAD model. 

This second variant is represented in Figure~\ref{fig:overview}-(b), and  defined as \ie $\hat{y} = f(\crop; \classw)$, where $\crop$ denotes the ROI. Independent ConvNets were trained to approximate $\class(\cdot)$ in the full image and ROI cases. The later was trained as well with a cross entropy-loss but using a ROI-only dataset.

\subsection{Classification after Automatic Localization} 
In this subsection, we focused on an automatic method to localize the ROI within the radiograph. We leveraged the bounding box annotations, manually provided by our experts, to formulate a secondary regression problem aiming to find the ROI in the radiograph. To this end, an auxiliary ConvNet was trained to predict the center and appropriate scale of the bounding box. 

We model the localization $g(\cdot)$ and classification $\class(\cdot)$ as independent tasks, as depicted in Figure~\ref{fig:overview}-(c).
The model for classification is equivalent to the one from the previous section trained on manual ROIs, while $g(\cdot)$ is modeled with a regression ConvNet minimizing the loss:
\begin{equation}
\mathcal{L}_{loc} = \frac{1}{2}\| \params - \hat{\params}\|^2, 
\label{eq:l2}
\end{equation}
where $\|\cdot\|$ is the $\ell_2$-norm, and $\hat{\params}$ is the predicted bounding box.
The output localized ROI image is then obtained as $\crop = \warp(\im)$ and fed to $f(\crop; \classw)$, only for evaluation. 

\subsection{Model Architectures and Implementation Details}
For classification tasks, we used a Residual Network (ResNet-50) \cite{He2016:resnet}, which was pretrained on ImageNet. The network was trained on radiographs, down-sampled from the original size to $224 \times 224$ px. In our case, the categories are the classes in the AO standard (type A and B) and not-fractured. Data augmentation techniques such as translation, scaling and rotation were used. The localization network was designed following AlexNet \cite{Krizhevsky2012:imagenet}. For this architecture, full X-ray images were down-sampled to $227 \times 227$ px. All the models were trained on a Linux based workstation equipped with 16GB RAM, Intel(R) Xeon(R) CPU @ 3.50GHz and 64 GB GeForce GTX 1080 graphics card. Stochastic Gradient Descent was used for optimization. All the models were trained until convergence (80 and 200 epochs for classification and localization, respectively). The batch size and momentum were kept constant as 64 and 0.9 for all three models. The learning rate was initialized to $1\times10^{-2}$ for the classification models, and to $1\times10^{-8}$ for the localization network, the decay varied among the different models. 

\begin{figure*}[t]
    \centering
    \includegraphics[width=1.\textwidth]{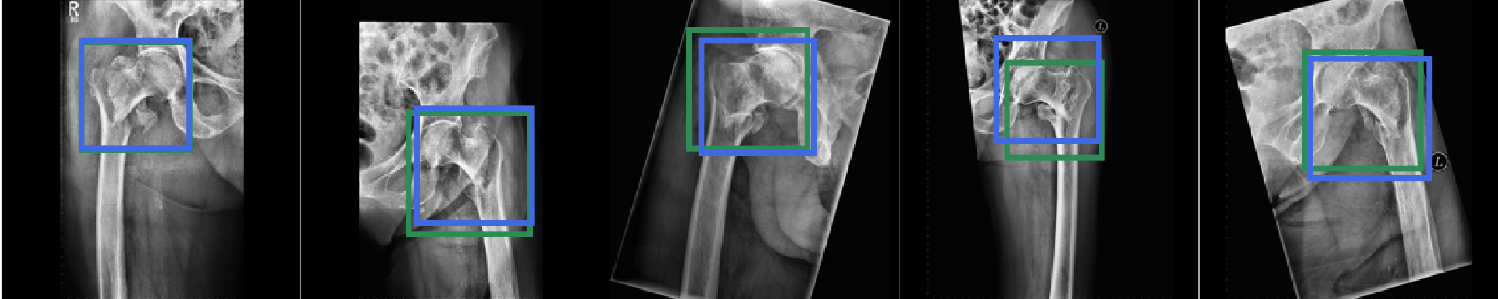}
        \caption{{\bf Localization capabilities of the CAD system.} Manually delineated (green) and predicted (blue) bounding boxes for the region of interest in the radiograph.}
    \label{fig:bboxes}
\end{figure*}

\section{Experimental Validation}
\paragraph{\textbf{Dataset Collection and Preparation}.}
X-rays of the hip and pelvis of 780 subsequently sampled patients (69\% female), with a mean patient age of 75.7 years $\pm$ 13.2, diagnosed with proximal femur fractures between 2007 and 2017 at the trauma surgery department of the Rechts der Isar Hospital in Munich were retrospectively gathered in an anonymized manner. The collected images of each patient contained either anterior-posterior (a-p) and lateral view (4\%) or only the a-p image. The anterior-posterior views of the pelvis with two hip joints and femora were parted into two, containing one femur each. In most cases, one of the images showed a normal, not-fractured contralateral femur. 

Regarding the classification, we looked at two scenarios: fracture detection (“not-fractured” vs. “abnormal”) and further division of the “abnormal” class into types “A” and “B”. Type “C” fractures were not included in the study as the number of cases was significantly lower than for the other classes. Such fractures are in fact more common in children and follow a different treatment path. For the two-class problem, 780 fracture images and 567 not-fractured images were considered. The same setting was used for the three-class problem considering 327-type A-, 453-type B-fractures and 567-not-fractured X-rays. The dataset was split patient-wise into three parts with the ratio 70\%: 10\%: 20\% to build respectively the training, validation and test set in all presented experiments. To train and test the CAD system, we collected class labels from three clinical experts: one trauma surgeon, one senior radiologist, and one 5th-year resident trauma surgeon (under the supervision of the radiologist). Each of them evaluated a split of the dataset. The test set was designed to have a class-balanced distribution between classes A, B and not-fractured, consisting of $(55, 60, 55)$ images, respectively. An additional set of 55 not-fractured images was included for the balanced comparison of the two-class scenario (not-fractured: 115 vs. abnormal: 115). 

\paragraph{\textbf{Evaluation Metrics}.}
We used standard classification metrics derived from the confusion matrices: accuracy, precision, recall and $F_1$-score, and the Area Under the Curve (AUC) for the Receiver Operating Characteristic (ROC) curve. To evaluate the localization precision of our network, we reported the percentage of ROI centers that are contained in the manually provided bounding boxes. The retrieval task was evaluated using the 11-point precision recall curve~\cite{Zhang2009}. 

\begin{table*}[t]
\centering
\ra{1.3}
\caption{{\bf Classification metrics for evaluation of the three compared methods: full radiographs, manually defined ROIs (Manual ROIs), and after automatically predicting the ROIs (Automatic Localization); and the average clinical expert.} Accuracy, precision, recall and $F_1$-score of our models. The highest metric values across the three models are highlighted in bold for each metric and classification type.}
\label{table:metrics}
\begin{tabular}{@{}rcccccc@{}}
\toprule
 & 2 classes & \phantom{a} & \multicolumn{4}{c}{3 classes} \\
 \cmidrule{2-2} \cmidrule{4-7} 
 & Abnormal && Type A & Type B & Normal & Avg.\\ \midrule
\textbf{Full Radiographs}\\ 
Accuracy & 83\% && 86\% & 87\% & \textbf{94\%} & 89\% \\
Precision & 78\% && 86\% & 78\% & 88\%  & 84\%  \\
Recall & 83\% && 67\% & 83\% & 95\% & 82\% \\
$F_1$-score & 84\% && 76\% & 82\% & \textbf{91\%} & 83\% \\
\midrule
\textbf{Manual ROIs}\\ 
Accuracy & \textbf{93}\% && \textbf{91\%} & \textbf{91\%} & 91\% & \textbf{91\%} \\
Precision & 93\% && \textbf{98\%} & \textbf{87\%} & 81\% & \textbf{88\%}  \\
Recall & \textbf{94\%} && \textbf{75\%} & 88\% & \textbf{97\%} & \textbf{87\%} \\
$F_1$-score & \textbf{94\%} && \textbf{85\%}  & \textbf{88\%}  & 88\% & \textbf{87\%} \\
\midrule
\textbf{Automatic Localization}\\ 
Accuracy & \textbf{93\%} && 89\% & 87\% & \textbf{94\%} & 90\% \\
Precision & \textbf{94\%} && 90\% & 77\% & 90\% & 86\%  \\
Recall & 93\% && 73\% & \textbf{90\%} & 92\% & 85\% \\
$F_1$-score & 93\% && 81\% & 83\% & \textbf{91\%} & 85\% \\
\midrule
\textbf{Clinical Expert}\\ 
Accuracy & 92\% && 92\% & 89\% & 93\% & \textbf{91\%} \\
Precision & 92\% && 90\% & 79\% & 94\% & \textbf{88\%}  \\
Recall & 92\% && 83\% & 92\% & 86\% & \textbf{87\%} \\
$F_1$-score & 92\% && 86\% & 95\% & 90\% & \textbf{87\%} \\
\bottomrule
 \end{tabular}
 \end{table*}

\section{Results}
\subsection{Classification on Full Radiographs}

Two classification scenarios of proximal femur fractures were evaluated. First, the two-class fracture detection, \ie differentiating between not-fractured and abnormal cases. Second, discriminating among three classes: not-fractured, type A- or type B-fracture. In Table~\ref{table:metrics}, we present the accuracy, precision recall and $F_1$-score for the classification of full X-ray images in two hierarchical scenarios. The performance of the model is maintained when the number of classes was increased from 2 to 3, with an average $F_1$-score of 84\% and 83\% respectively. 

\subsection{Classification on Manual ROIs}
As it can be seen in Table 1, the use of a region of interest instead of the full image increased all the classification metrics, showing the importance of localization as reported in \cite{kazi2017}. In this case, the network visualizes the characteristics (location, shape, number of fragments) of the fracture at a preferable resolution. $F_1$-score improvement accounts for 12\% (from 0.84 to 0.94) for fracture detection. When the model has to differentiate between fracture type “A” and “B”, due to the increased difficulty of the task,
the improvement is accounted for 5\%. Previous work was reported for only fracture detection, \ie binary classification, our results with an AUC of 0.98 are comparable to state-of-the-art~\cite{Cheng2019,Wang2019WeaklySupervised}. In addition, an AUC of 0.95 was obtained for the three-class problem.

\subsection{Classification after Automatic Localization}
In Figure~\ref{fig:bboxes}, some examples of the manually provided and the predicted bounding boxes are illustrated. Even though there is not a unique way to define a bounding box, we found that the manually defined bounding boxes always (100\%) contained the center of the predicted ROIs. Based on the metrics reported in Table~\ref{table:metrics}, we observed that the classification model performed similarly on
the automatically extracted regions ($F_1$-score of 93\%) and the ones manually provided by the experts ($F_1$-score of 94\%). However, the automatic localization removes the need of an expert intervention during test time.

\begin{figure*}[]
    \centering
    \includegraphics[width=1.\textwidth]{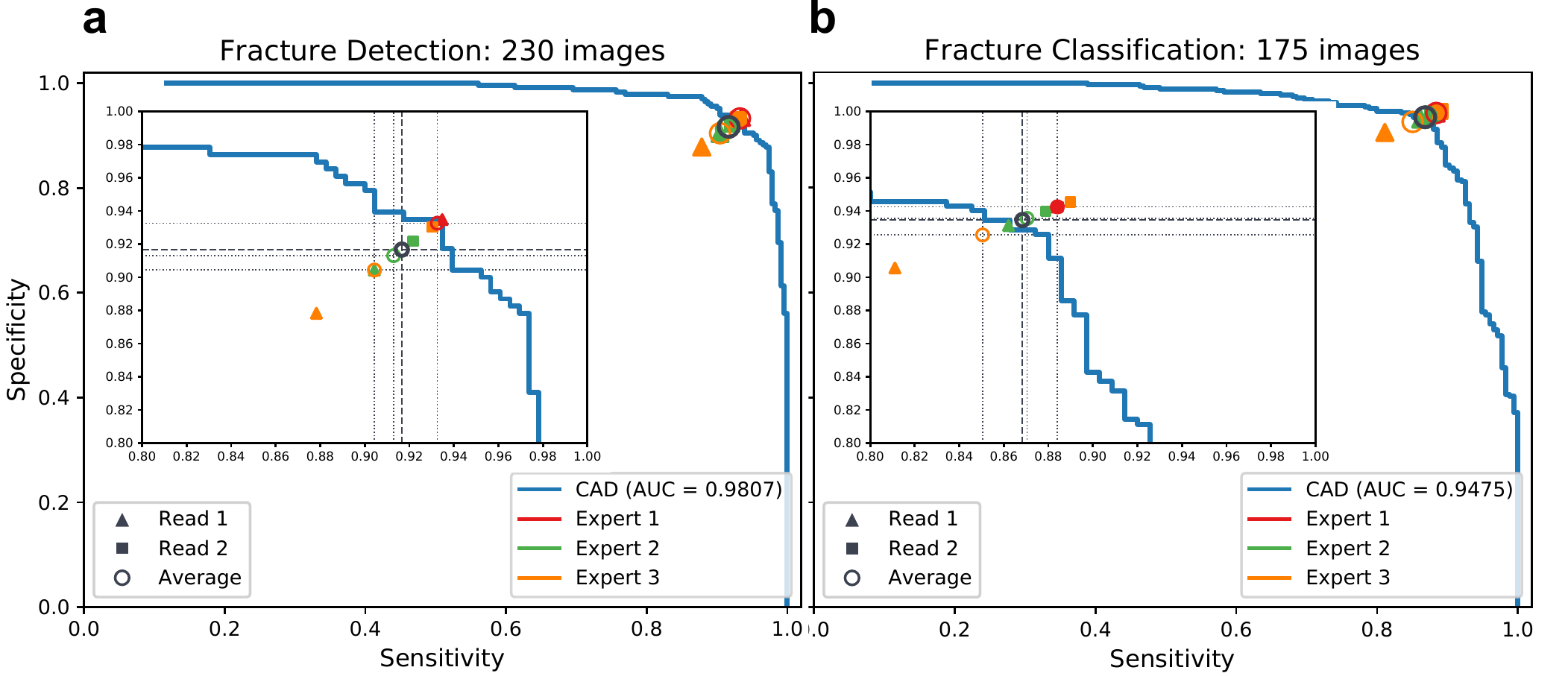}
        \caption{{\bf Clinical experts and CAD performance.} Comparison of specificity measuring the proportion of negatives correctly identified, against sensitivity accounting for the number of positives correctly found, for (a) fracture detection and (b) classification. The set of colors distinguish the CAD system from the individual experts. The filled shapes illustrate the first (triangle) and second (square) readings. The mean performance of every expert is depicted by colored circles, and the average clinical expert by a black circle. } 
    \label{fig:rocs} 
\end{figure*}

\begin{figure*}[]
    \centering
    \includegraphics[width=0.8\textwidth]{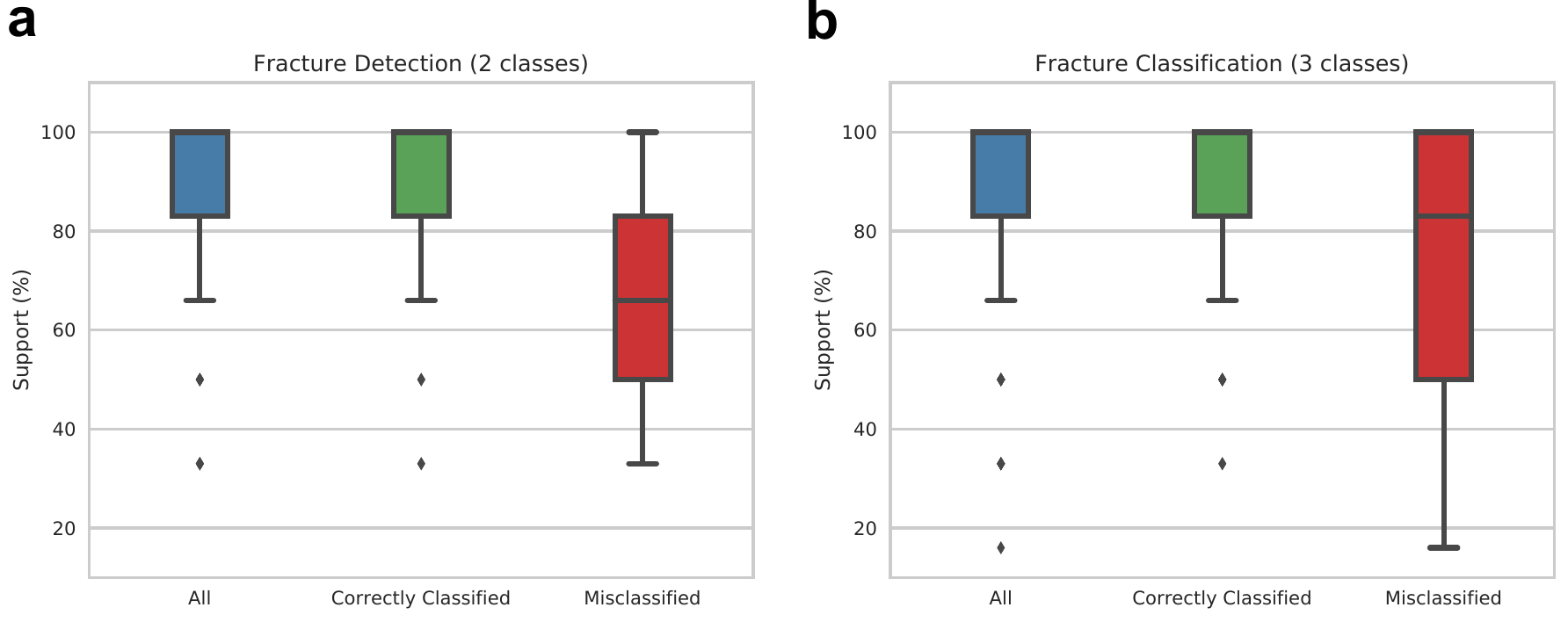}
        \caption{{\bf  Classification robustness and informative disagreement across scales.} Percentage of agreeing predictions across different input scales: $[0.75, 1.00, 1.25, 1.50, 1.75, 2.00]$. We gathered the predictions of the scaled regions of interest, and quantified the number of correctly classified for {\bf (a)} fracture detection and {\bf (b)} classification. The boxplot shows the median and standard deviation of the support for all test images (in blue), correctly classified (in green) and misclassified (in red). } 
    \label{fig:boxplot} 
\end{figure*}

\subsection{Expert-level Performance}
In order to evaluate the relevance of the previously obtained results, we compared the best-performing model against the individual performance of three clinical experts. We asked three experts (a trauma surgery attendant, a senior radiologist and a trauma surgery attending 1st year) to read the test-set twice with a 2 to 3-week interval between the readings. The evaluation of this set of images took on average 46 minutes. In Table~\ref{table:metrics}, the average performance of the experts in the two readings is reported. A ROC analysis was carried out by using the ground truth as target classification. To this end, ROC curves were built from the reciprocal relation between sensitivity and specificity calculated for all the possible threshold values. In Figure~\ref{fig:rocs}, only for visualization purposes, the x-axis has been inverted, \ie we show “Sensitivity” instead of “1-Sensitivity”. The average performance between the two readings of each expert compared to the others can be analyzed, but in this case, only one point of the ROC space is obtained. According to the metrics in Table~\ref{table:metrics} and Figure~\ref{fig:rocs}, our CAD model, trained on manual ROIs, performed similarly compared to the average expert results in fracture classification, and it performed even better regarding the binary fracture detection task. 

\subsection{Robustness and Retrieval}
\textbf{Scale sensitiveness analysis.}
We further investigated the robustness of our model against the variability of the scale of the predicted ROIs. The predicted bounding boxes were scaled by the following values [0.75,1.00,1.25,1.50,1.75,2.00] and fed to the classification network. We gathered the predictions at each scale and quantified the percentage of correct predictions across scales, i.e. the number of scaled ROIs supporting the correct classification. Results are reported in Figure~\ref{fig:boxplot}. They show a mean support for the correct prediction of 93.82\% and 88.35\% with 12.34\% and 16.58\% standard deviation for 2 and 3 classes, respectively. These values demonstrate the robustness and stability of the CAD system to scale variations for most of the cases. Moreover, the disagreement across different scales was shown to be informative of spurious predictions, as suggested by \cite{Cheplygina2018:disagreement}.

\begin{figure*}[]
    \centering
    \includegraphics[width=0.9\textwidth]{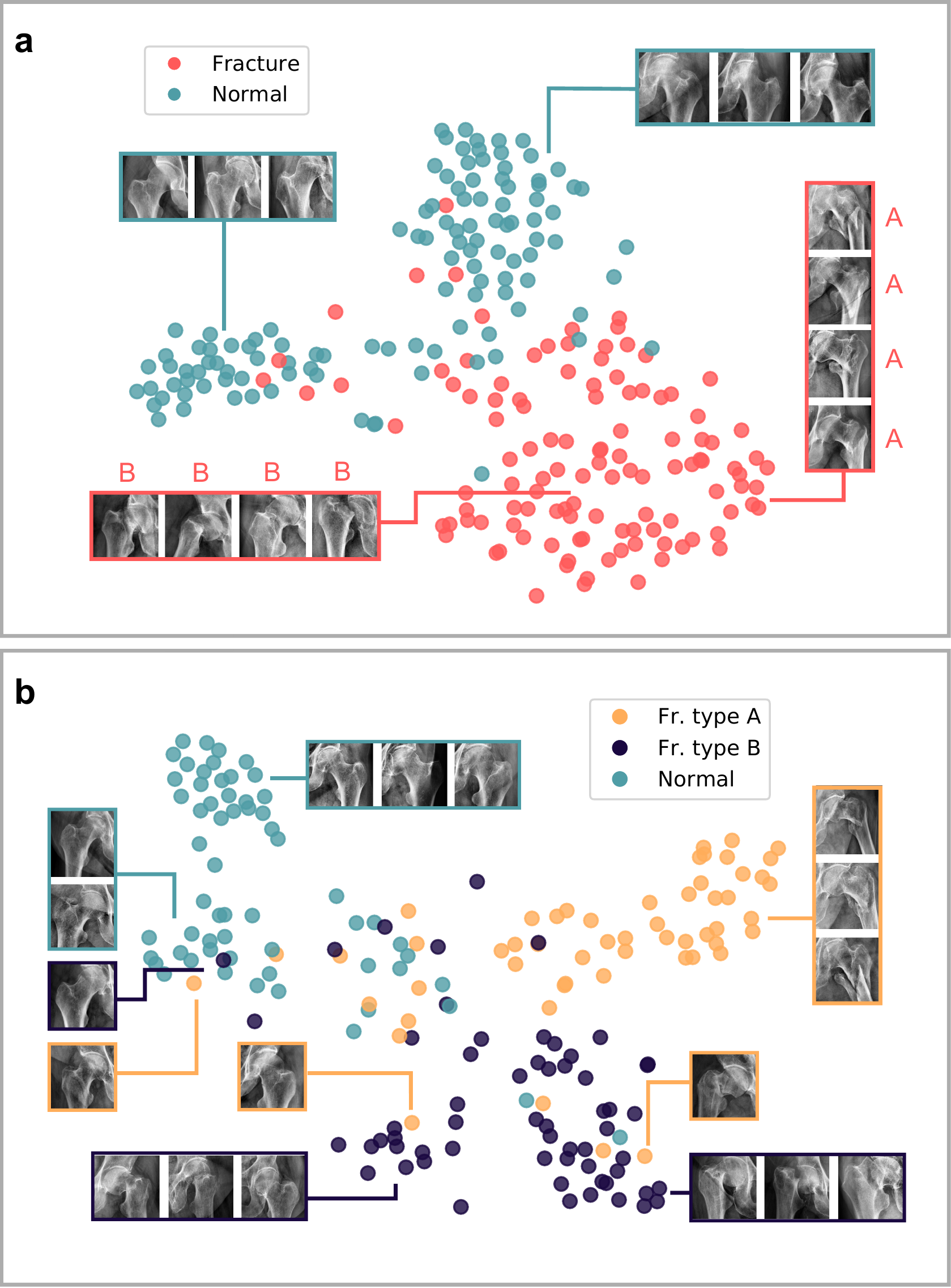}
        \caption{{\bf Projected 2D space learned by t-SNE.} {\bf (a)} Fracture detection and {\bf (b)} Fracture classification. At the top part, we observe that the model was able to differentiate and group left or right femur. These two clusters were especially differentiated in the not-fractured (``normal'') class. Moreover, within the abnormal examples, images of type A and B were differentiated, even if the network was only trained for binary classification.
        }
    \label{fig:tsne}
\end{figure*}

\begin{figure*}[t]
    \centering
    \includegraphics[width=0.6\textwidth]{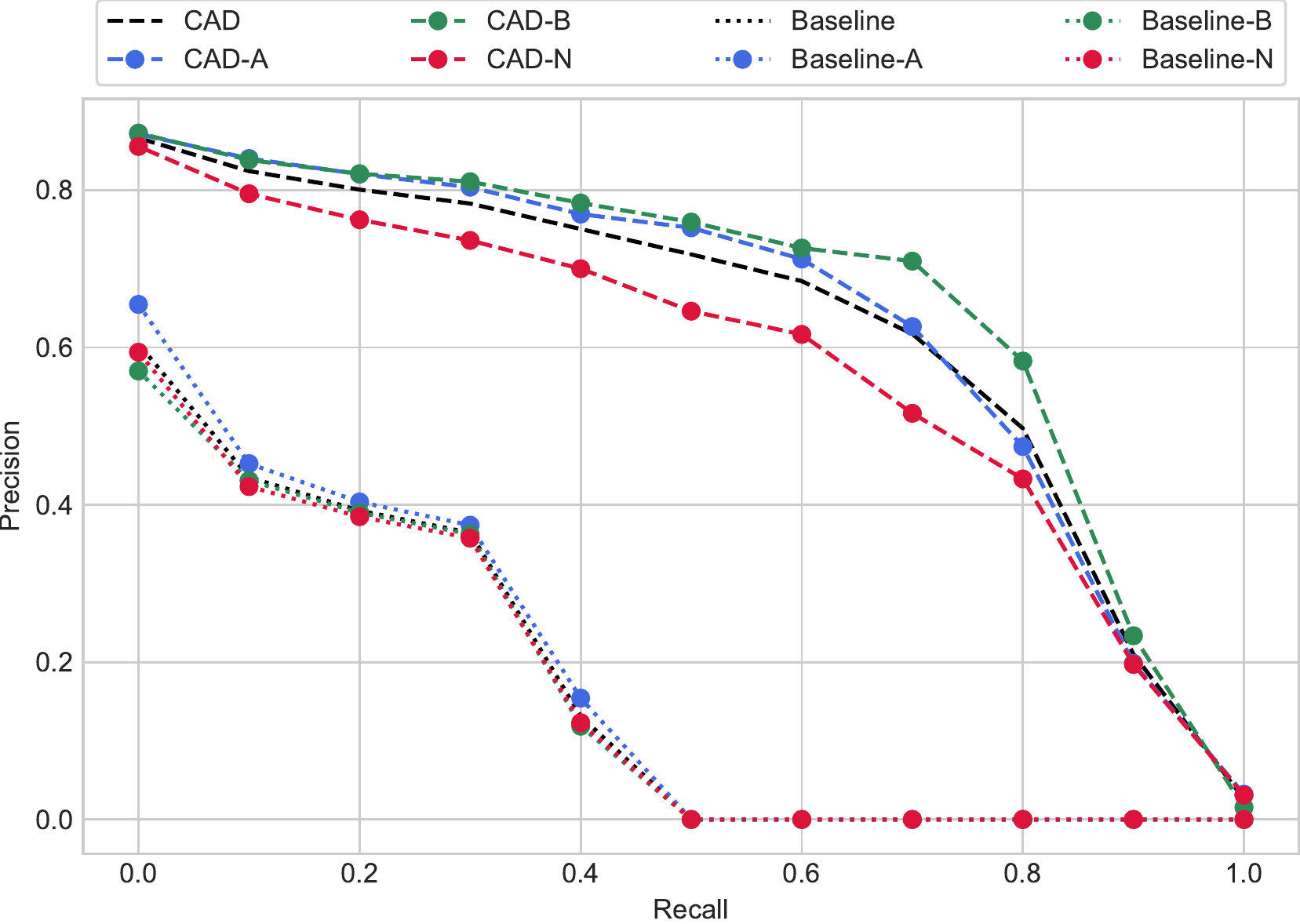}
        \caption{{\bf Precision \textit{vs}. recall in the image retrieval task.}  The dashed line represents our best-performing CAD model and the dotted line the baseline. In black color is depicted the average performance, while the colors stand for each of the classes. } 
    \label{fig:retrieval} 
\end{figure*}

\textbf{Clinical use case: image retrieval.} Here we use the penultimate layer of the network to produce a lower-dimensional representation of each image. We then measure the distance between an unseen query image (from the test set) and the pool of retrievable images (corresponding to the training set), in order to retrieve the most similar cases to the query. We verify the relevance of the retrieval system, by projecting the learned feature representation of the testing images to two dimensions by means of the t-SNE algorithm \cite{maaten2008visualizing}. In the embedded space, depicted in Figure \ref{fig:tsne}, the points belonging to different classes for both the 2- and 3-class problems are successfully separated. We evaluate the classification model trained on manual ROIs for retrieval in terms of precision and recall. The precision measures the proportion of relevant images (of the same class as the query) among the retrieved ones. On average, when retrieving 10 images, 9 of the proposed results are relevant. The recall evaluates how many relevant images are retrieved out of the total number of relevant cases. On average, with 100 retrieved images, we recover almost 70\% of the relevant cases in the training set. We summarize the results for different numbers of retrieved images $[5,10,30,50,80,100,200,300,400]$ in the 11-point precision-recall curve~\cite{Zhang2009} in Figure~\ref{fig:retrieval}. This curve is based on the Euclidean distances between the query and retrievable images. We compared our CAD model, where the distances are computed on the CNN embedded space, against the distances of the ``raw'' images (baseline). The proposed CAD retrieval reaches a mean average precision of 0.62 compared to 0.18 of the baseline.

\section{Discussion}
Proximal femur fractures present a huge socioeconomic problem especially in the elderly population. The adequate and exact graduation of these fractures according to the AO classification is highly important for the following treatment and the clinical outcome of the patients. Different to any prior approach, the presented work focuses on a method capable of identifying multiple classes according to a clinical classification standard such as the AO. Our CAD framework exhibits a high $F_1$-score and AUC of 94\% and 0.98 respectively for the two-class problem when differentiating between fracture \vs not-fractured, and of 87\% and 0.95  for the three-class problem when not-fractured is further divided into type A- and B-fracture. These classification metrics are comparable to state-of-the-art results \cite{Cheng2019,Wang2019WeaklySupervised}. These high values indicate that our CAD is suitable to be implemented in the daily clinical routing of trauma surgeons treating proximal femur fractures.

\paragraph{\textbf{Clinical impact.}} In the short term, our system may assist the trauma surgery residents during their daily clinical training by profiting of a second reading from our CAD tool. In addition, the use of retrieval cases provides an opportunity to focus training through the query of similar and ambiguous cases. Since an inadequate initial classification may lead to an inappropriate treatment plan, it could support residents in trauma surgery, especially in small peripheral hospitals, to reach a more adequate decision. The most impactful application of such a fracture classification tool would be in the everyday surgery planning, where it is likely that a CAD system could help in reducing fatigue while improving accuracy, given that such a system cannot be affected by bias, experience or workload. In the trauma surgery department, often clinicians are faced with emergency situations and decisions have to be made fast. Due to the fact that our method is able to classify on the fly X-ray images, it could also assist in triaging patients in the emergency room. In cases of proximal femur fractures, time until diagnosis is critical. In fact early surgery and mobilization have been identified as key factors in reducing the number of complications after surgery and mortality \cite{Flikweert2018}. In treating fracture neck femurs (B-type fractures), classification remains the best option to reduce the risk of complications like non-union and avascular necrosis in treating fracture neck femurs \cite{Mittal2012}. In this context, a waiting time of 24 hours until surgery have been shown to be associated with a greater risk of 30-day mortality \cite{Pincus2017}. 

\begin{figure*}[t]
    \centering
    \includegraphics[width=1.0\textwidth]{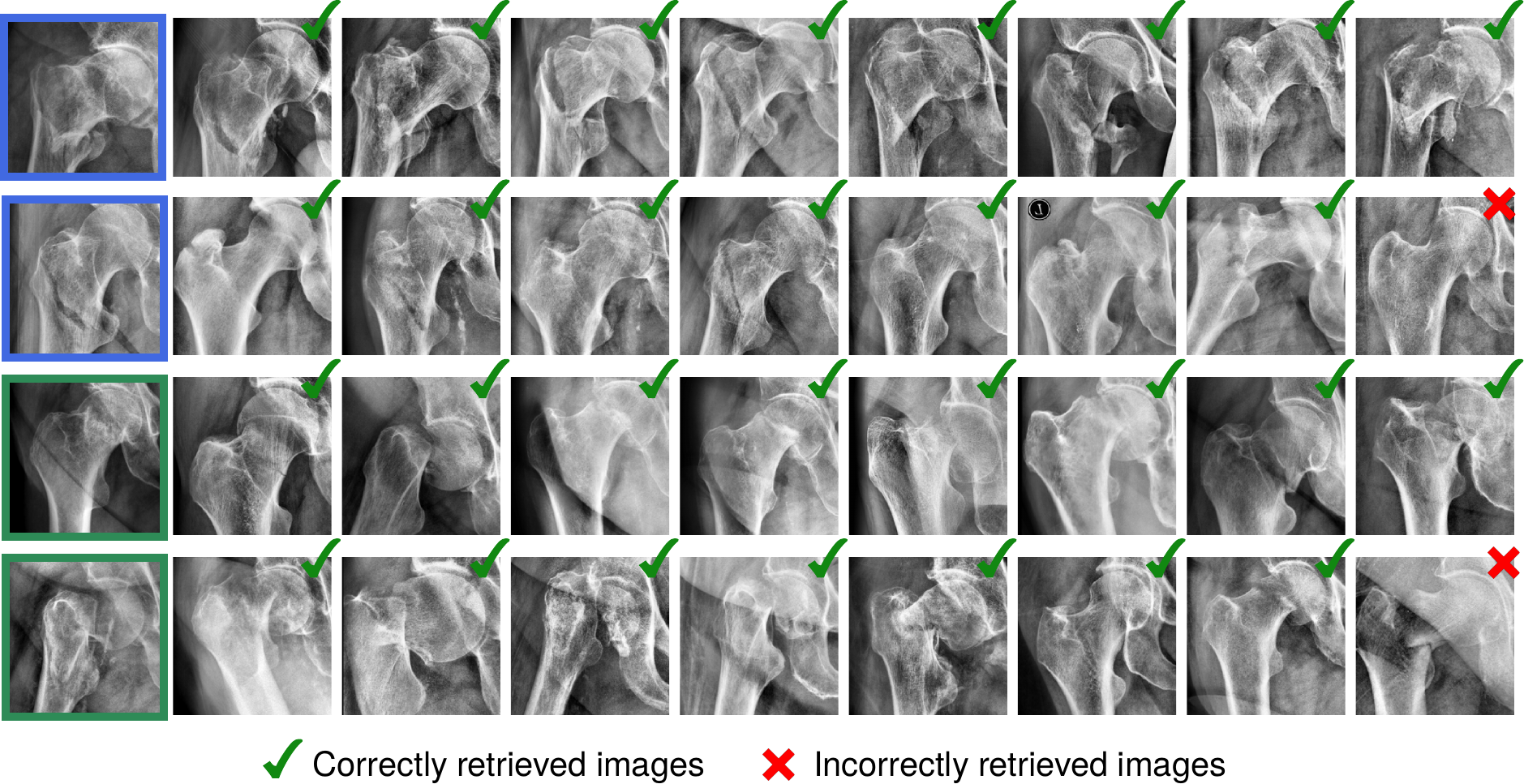}
    \caption{{\bf Query and retrieval examples.} Query images are surrounded by a colored box: A-type fracture (blue) and B-type (green) fracture. For each query, the closest 8 retrieved images are shown. On average, when retrieving 10 images with our CAD model, 9 of the proposed results present the correct classification.}
    \label{fig:queries}
\end{figure*}

\paragraph{\textbf{Adoption into clinical practice.}} In order to favor the integration of the proposed CAD system into the daily clinical routine, we propose three layers of verification. First, along with the predicted class labels, we provide as an additional output the localization of the fracture in the form of a bounding box. A proper working system provides a bounding box leaving out any irrelevant regions. If it were the case that the system predicts an area of support outside the expected anatomy, a supportive minimal interaction tool is procured, consisting of two clicks to manually select the ROI around the fracture. Classification on the selected ROI leads to both a positive and significant improvement in the classification as described in Results section. A second verification is the agreement of the class predictions over several scales (see Figure \ref{fig:boxplot}), where we found that disagreement could point out misclassification examples. Finally, we use the learned feature space of the network to retrieve similar images (see Figure \ref{fig:queries}). Such a retrieval system may be used by residents to learn variants of a single fracture type or even useful for experts to analyze complex cases in comparison with the retrieved samples.  In this way, our method helps to speed up the diagnosis and treatment planning for complex fractures cases.

\paragraph{\textbf{Technical limitations.}}
From the technical side, our dataset suffers from a high imbalance in the distribution of the classes when considering the subtypes of A- and B- fractures. Such scenarios require more complex (and thus less interpretable models) \cite{kazi2017} or different optimization strategies \cite{Bengio2009Curriculum,Jimenez2019MedicalCurriculum}. We further plan to investigate weighting schemes, for example based on the uncertainty of the model \cite{Gal2016BayesianDropout}, or relying on triplet metric learning~\cite{huang2016learning}. Exploring bootstrapping strategies could be one way of handling noisy labels \cite{reed2014training}. One could also consider modeling label's uncertainty to estimate the uncertainty due to intra or inter-observer variability \cite{Tomczack2019LearnTE}. 

\section{Conclusions}
We have proposed a CAD scheme for the detection and further classification of proximal femur fractures achieving results comparable to state-of-the-art performance for the binary fracture detection task. Moreover, we show for the first time, an in-depth evaluation of an automatic system for the multi-class problem according to the AO system. This level of categorization is crucial for planning the treatment either conservatively or surgically, and if so, to choose the adequate surgical implant. The localization of the region of interest was highly accurate, all the predicted centers of the ROI were contained in the original bounding box. The sensitivity of the system to the size of the ROI was analyzed in detail; we found that disagreement in classification at different ROI sizes could signal the potential for misclassification. We presented a clinical use case of retrieval to assist the training of trauma surgery residents, especially for those working in small peripheral hospitals. Finally, we discussed several strategies of verification to favor the adoption of our CAD tool into the daily clinical routine.

\section*{Funding}
 A. Jim\'{e}nez-S\'{a}nchez has received financial support through the ``la Caixa'' Foundation (ID Q5850017D), fellowship code: LCF/BQ/IN17/11620013, and from the European Union’s Horizon 2020 research and innovation programme under the Marie Sk\l{}odowska-Curie grant agreement No. 713673. A. Kazi is financially supported by Freunde und F{\"o}rderer der Augenklinik, M{\"u}nchen, Germany. D. Mateus has received funding from Nantes M\'etropole and the European Regional Development, Pays de la Loire, under the Connect Talent scheme. SA is supported by the PRIME programme of the German Academic Exchange Service (DAAD) with funds from the German Federal Ministry of Education and Research (BMBF).

\section*{Compliance with Ethical Standards}
This projected has received approval of the ethics committee under the number: 409/15 S.

\section*{Informed consent}
There is a general agreement in the University Hospital Rechts der Isar in Munich (Germany), that whenever a patient signs for X-ray images to be taken, these images might be used for scientific studies in a complete anonymized way as described in Dataset Collection and Preparation subsection. 

\section*{Conflict of interest}
The authors declare that they have no conflict of interest.

\begin{acknowledgements}
The authors would like to thank our clinical partners, in particular, Ali Deeb, PD.  Dr. med. Marc Beirer, and Fritz Seidl, MA, MBA. for their support during the work. The authors would like to thank Nvidia for the donation of a GPU.
\end{acknowledgements}

\bibliographystyle{ieeetr}
\bibliography{biblio.bib}

\begin{thebibliography}{10}

\bibitem{Ryan2015}
D.~J. Ryan, H.~Yoshihara, D.~Yoneoka, K.~A. Egol, and J.~D. Zuckerman, ``Delay
  in hip fracture surgery: An analysis of patient-specific and
  hospital-specific risk factors,'' {\em Journal of Orthopaedic Trauma},
  vol.~29, pp.~343--348, aug 2015.

\bibitem{Giannoulis2016}
D.~Giannoulis, G.~M. Calori, and P.~V. Giannoudis, ``Thirty-day mortality after
  hip fractures: has anything changed?,'' {\em European Journal of Orthopaedic
  Surgery {\&} Traumatology}, vol.~26, pp.~365--370, mar 2016.

\bibitem{DeBellis2014}
U.~G.~D. Bellis, C.~Legnani, and G.~M. Calori, ``Acute total hip replacement
  for acetabular fractures: A systematic review of the literature,'' {\em
  Injury}, vol.~45, pp.~356--361, feb 2014.

\bibitem{Sheehan2015}
S.~E. Sheehan, J.~Y. Shyu, M.~J. Weaver, A.~D. Sodickson, and B.~Khurana,
  ``Proximal femoral fractures: What the orthopedic surgeon wants to know,''
  {\em {RadioGraphics}}, vol.~35, pp.~1563--1584, Sept. 2015.

\bibitem{Bhandari2015}
M.~Bhandari, P.~J. Devereaux, T.~A. Einhorn, L.~Thabane, E.~H. Schemitsch,
  K.~J. Koval, F.~Frihagen, R.~W. Poolman, K.~Tetsworth, E.~Guerra-Farfan,
  K.~Madden, S.~Sprague, G.~Guyatt, and H.~Investigators, ``Hip fracture
  evaluation with alternatives of total hip arthroplasty versus
  hemiarthroplasty ({HEALTH}): protocol for a multicentre randomised trial,''
  {\em {BMJ} Open}, vol.~5, pp.~e006263--e006263, feb 2015.

\bibitem{Zuckerman1996hip}
J.~D. Zuckerman, ``Hip fracture,'' {\em New England journal of medicine},
  vol.~334, no.~23, pp.~1519--1525, 1996.

\bibitem{Grigoryan2014ortho}
K.~V. Grigoryan, H.~Javedan, and J.~L. Rudolph, ``Ortho-geriatric care models
  and outcomes in hip fracture patients: a systematic review and
  meta-analysis,'' {\em Journal of orthopaedic trauma}, vol.~28, no.~3, p.~e49,
  2014.

\bibitem{Swiontkowski1984fractures}
M.~F. Swiontkowski, R.~Winquist, and J.~S. Hansen, ``Fractures of the femoral
  neck in patients between the ages of twelve and forty-nine years.,'' {\em The
  Journal of bone and joint surgery. American volume}, vol.~66, no.~6,
  pp.~837--846, 1984.

\bibitem{Parkkari1999majority}
J.~Parkkari, P.~Kannus, M.~Palvanen, A.~Natri, J.~Vainio, H.~Aho, I.~Vuori, and
  M.~J{\"a}rvinen, ``Majority of hip fractures occur as a result of a fall and
  impact on the greater trochanter of the femur: a prospective controlled hip
  fracture study with 206 consecutive patients,'' {\em Calcified tissue
  international}, vol.~65, no.~3, pp.~183--187, 1999.

\bibitem{Kellam2018}
J.~F. Kellam, E.~G. Meinberg, J.~Agel, M.~D. Karam, and C.~S. Roberts,
  ``Introduction,'' {\em Journal of Orthopaedic Trauma}, vol.~32, pp.~S1--S10,
  jan 2018.

\bibitem{Jin2005}
W.-J. Jin, L.-Y. Dai, Y.-M. Cui, Q.~Zhou, L.-S. Jiang, and H.~Lu, ``Reliability
  of classification systems for intertrochanteric fractures of the proximal
  femur in experienced orthopaedic surgeons,'' {\em Injury}, vol.~36,
  pp.~858--861, jul 2005.

\bibitem{vanEmbden2010}
D.~van Embden, S.~Rhemrev, S.~Meylaerts, and G.~Roukema, ``The comparison of
  two classifications for trochanteric femur fractures: The {AO}/{ASIF}
  classification and the jensen classification,'' {\em Injury}, vol.~41,
  pp.~377--381, apr 2010.

\bibitem{wu2012fracture}
J.~Wu, P.~Davuluri, K.~R. Ward, C.~Cockrell, R.~Hobson, and K.~Najarian,
  ``Fracture detection in traumatic pelvic ct images,'' {\em Journal of
  Biomedical Imaging}, vol.~2012, p.~1, 2012.

\bibitem{al2013detecting}
M.~Al-Ayyoub, I.~Hmeidi, and H.~Rababah, ``Detecting hand bone fractures in
  x-ray images.,'' {\em JMPT}, vol.~4, no.~3, pp.~155--168, 2013.

\bibitem{bayram:2016diffract}
F.~Bayram and M.~{\c{C}}ak{\i}ro{\u{g}}lu, ``Diffract: Diaphyseal femur
  fracture classifier system,'' {\em Biocybernetics and Biomedical
  Engineering}, vol.~36, no.~1, pp.~157--171, 2016.

\bibitem{roth2016deep}
H.~R. Roth, Y.~Wang, J.~Yao, L.~Lu, J.~E. Burns, and R.~M. Summers, ``Deep
  convolutional networks for automated detection of posterior-element fractures
  on spine ct,'' in {\em Medical Imaging 2016: Computer-Aided Diagnosis},
  vol.~9785, p.~97850P, International Society for Optics and Photonics, 2016.

\bibitem{olczak2017artificial}
J.~Olczak, N.~Fahlberg, A.~Maki, A.~S. Razavian, A.~Jilert, A.~Stark,
  O.~Sk{\"o}ldenberg, and M.~Gordon, ``Artificial intelligence for analyzing
  orthopedic trauma radiographs: deep learning algorithms—are they on par
  with humans for diagnosing fractures?,'' {\em Acta orthopaedica}, vol.~88,
  no.~6, pp.~581--586, 2017.

\bibitem{Kitamura2019}
G.~Kitamura, C.~Y. Chung, and B.~E. Moore, ``Ankle fracture detection utilizing
  a convolutional neural network ensemble implemented with a small sample, de
  novo training, and multiview incorporation,'' {\em Journal of Digital
  Imaging}, vol.~32, pp.~672--677, Aug 2019.

\bibitem{Urakawa2019}
T.~Urakawa, Y.~Tanaka, S.~Goto, H.~Matsuzawa, K.~Watanabe, and N.~Endo,
  ``Detecting intertrochanteric hip fractures with orthopedist-level accuracy
  using a deep convolutional neural network,'' {\em Skeletal Radiology},
  vol.~48, pp.~239--244, Feb 2019.

\bibitem{Badgeley2019}
M.~A. Badgeley, J.~R. Zech, L.~Oakden-Rayner, B.~S. Glicksberg, M.~Liu,
  W.~Gale, M.~V. McConnell, B.~Percha, T.~M. Snyder, and J.~T. Dudley, ``Deep
  learning predicts hip fracture using confounding patient and healthcare
  variables,'' {\em npj Digital Medicine}, vol.~2, Apr. 2019.

\bibitem{Cheng2019}
C.-T. Cheng, T.-Y. Ho, T.-Y. Lee, C.-C. Chang, C.-C. Chou, C.-C. Chen, I.-F.
  Chung, and C.-H. Liao, ``Application of a deep learning algorithm for
  detection and visualization of hip fractures on plain pelvic radiographs,''
  {\em European Radiology}, vol.~29, pp.~5469--5477, Oct 2019.

\bibitem{Wang2019WeaklySupervised}
Y.~Wang, L.~Lu, C.-T. Cheng, D.~Jin, A.~P. Harrison, J.~Xiao, C.-H. Liao, and
  S.~Miao, ``Weakly supervised universal fracture detection in pelvic x-rays,''
  in {\em Medical Image Computing and Computer Assisted Intervention -- MICCAI
  2019} (D.~Shen, T.~Liu, T.~M. Peters, L.~H. Staib, C.~Essert, S.~Zhou, P.-T.
  Yap, and A.~Khan, eds.), (Cham), pp.~459--467, Springer International
  Publishing, 2019.

\bibitem{Esteva2017}
A.~Esteva, B.~Kuprel, R.~A. Novoa, J.~Ko, S.~M. Swetter, H.~M. Blau, and
  S.~Thrun, ``Dermatologist-level classification of skin cancer with deep
  neural networks,'' {\em Nature}, vol.~542, pp.~115--118, jan 2017.

\bibitem{kazi2017}
A.~Kazi, S.~Albarqouni, A.~J. Sanchez, S.~Kirchhoff, P.~Biberthaler, N.~Navab,
  and D.~Mateus, ``Automatic classification of proximal femur fractures based
  on attention models,'' in {\em Machine Learning in Medical Imaging} (Q.~Wang,
  Y.~Shi, H.-I. Suk, and K.~Suzuki, eds.), (Cham), pp.~70--78, Springer
  International Publishing, 2017.

\bibitem{He2016:resnet}
K.~He, X.~Zhang, S.~Ren, and J.~Sun, ``Deep residual learning for image
  recognition,'' {\em 2016 IEEE Conference on Computer Vision and Pattern
  Recognition (CVPR)}, pp.~770--778, 2016.

\bibitem{Krizhevsky2012:imagenet}
A.~Krizhevsky, I.~Sutskever, and G.~E. Hinton, ``Imagenet classification with
  deep convolutional neural networks,'' in {\em Proceedings of the 25th
  International Conference on Neural Information Processing Systems - Volume
  1}, NIPS'12, (USA), pp.~1097--1105, Curran Associates Inc., 2012.

\bibitem{Zhang2009}
E.~Zhang and Y.~Zhang, {\em Eleven Point Precision-recall Curve}, pp.~981--982.
\newblock Boston, MA: Springer US, 2009.

\bibitem{Cheplygina2018:disagreement}
V.~Cheplygina and J.~P.~W. Pluim, ``Crowd disagreement about medical images is
  informative,'' in {\em Intravascular Imaging and Computer Assisted Stenting
  and Large-Scale Annotation of Biomedical Data and Expert Label Synthesis}
  (D.~Stoyanov, Z.~Taylor, S.~Balocco, R.~Sznitman, A.~Martel, L.~Maier-Hein,
  L.~Duong, G.~Zahnd, S.~Demirci, S.~Albarqouni, S.-L. Lee, S.~Moriconi,
  V.~Cheplygina, D.~Mateus, E.~Trucco, E.~Granger, and P.~Jannin, eds.),
  (Cham), pp.~105--111, Springer International Publishing, 2018.

\bibitem{maaten2008visualizing}
L.~v.~d. Maaten and G.~Hinton, ``Visualizing data using t-sne,'' {\em Journal
  of machine learning research}, vol.~9, no.~Nov, pp.~2579--2605, 2008.

\bibitem{Flikweert2018}
E.~R. Flikweert, K.~W. Wendt, R.~L. Diercks, G.~J. Izaks, D.~Landsheer,
  M.~Stevens, and I.~H.~F. Reininga, ``Complications after hip fracture
  surgery: are they preventable?,'' {\em European journal of trauma and
  emergency surgery : official publication of the European Trauma Society},
  vol.~44, pp.~573--580, Aug 2018.
\newblock 28795198[pmid].

\bibitem{Mittal2012}
R.~Mittal and S.~Banerjee, ``Proximal femoral fractures: Principles of
  management and review of literature,'' {\em Journal of Clinical Orthopaedics
  and Trauma}, vol.~3, pp.~15--23, June 2012.

\bibitem{Pincus2017}
D.~Pincus, B.~Ravi, D.~Wasserstein, A.~Huang, J.~M. Paterson, A.~B. Nathens,
  H.~J. Kreder, R.~J. Jenkinson, and W.~P. Wodchis, ``{Association Between Wait
  Time and 30-Day Mortality in Adults Undergoing Hip Fracture Surgery},'' {\em
  JAMA}, vol.~318, pp.~1994--2003, 11 2017.

\bibitem{Bengio2009Curriculum}
Y.~Bengio, J.~Louradour, R.~Collobert, and J.~Weston, ``Curriculum learning,''
  in {\em Proceedings of the 26th Annual International Conference on Machine
  Learning}, ICML '09, (New York, NY, USA), pp.~41--48, ACM, 2009.

\bibitem{Jimenez2019MedicalCurriculum}
A.~Jim{\'e}nez-S{\'a}nchez, D.~Mateus, S.~Kirchhoff, C.~Kirchhoff,
  P.~Biberthaler, N.~Navab, M.~A. Gonz{\'a}lez~Ballester, and G.~Piella,
  ``Medical-based deep curriculum learning for improved fracture
  classification,'' in {\em Medical Image Computing and Computer Assisted
  Intervention -- MICCAI 2019} (D.~Shen, T.~Liu, T.~M. Peters, L.~H. Staib,
  C.~Essert, S.~Zhou, P.-T. Yap, and A.~Khan, eds.), (Cham), pp.~694--702,
  Springer International Publishing, 2019.

\bibitem{Gal2016BayesianDropout}
Y.~Gal and Z.~Ghahramani, ``Dropout as a bayesian approximation: Representing
  model uncertainty in deep learning,'' in {\em Proceedings of The 33rd
  International Conference on Machine Learning} (M.~F. Balcan and K.~Q.
  Weinberger, eds.), vol.~48 of {\em Proceedings of Machine Learning Research},
  (New York, New York, USA), pp.~1050--1059, PMLR, 20--22 Jun 2016.

\bibitem{huang2016learning}
C.~Huang, Y.~Li, C.~Change~Loy, and X.~Tang, ``Learning deep representation for
  imbalanced classification,'' in {\em Proceedings of the IEEE Conference on
  Computer Vision and Pattern Recognition}, pp.~5375--5384, 2016.

\bibitem{reed2014training}
S.~E. Reed, H.~Lee, D.~Anguelov, C.~Szegedy, D.~Erhan, and A.~Rabinovich,
  ``Training deep neural networks on noisy labels with bootstrapping,'' in {\em
  3rd International Conference on Learning Representations, {ICLR} 2015, San
  Diego, CA, USA, May 7-9, 2015, Workshop Track Proceedings} (Y.~Bengio and
  Y.~LeCun, eds.), 2015.

\bibitem{Tomczack2019LearnTE}
A.~Tomczack, N.~Navab, and S.~Albarqouni, ``Learn to estimate labels
  uncertainty for quality assurance,'' {\em ArXiv}, vol.~abs/1909.08058, 2019.

\end{thebibliography}

\end{document}